\documentclass{article}

\usepackage[preprint]{casml_2024}
\usepackage{natbib}
\usepackage[utf8]{inputenc} 
\usepackage[T1]{fontenc}    
\usepackage{hyperref}       
\usepackage{url}            
\usepackage{booktabs}       
\usepackage{amsfonts}       
\usepackage{nicefrac}       
\usepackage{microtype}      
\usepackage{xcolor}         
\usepackage[pdftex]{graphicx}
\usepackage{multicol}

\title{A Numerical Study of Chaotic Dynamics of K-S Equation with FNOs}

\author{%
  Surbhi Khetrapal\thanks{Corresponding author: surbhikhetrapal@uohyd.ac.in} \\
  School of Physics,
  University of Hyderabad\\
  Hyderabad, Telangana, India.  \\
  \texttt{surbhikhetrapal@uohyd.ac.in} \\
   \And
   Jaswin Kasi \\
  Mercedes-Benz Research and Development\\
 Bengaluru, Karnataka, India. \\
  \texttt{
kasi.jaswin@mercedes-benz.com} \\
}

\begin{document}

\maketitle

\begin{abstract}
Solving non-linear partial differential equations which exhibit chaotic dynamics is an important problem with a wide-range of applications such as predicting weather extremes and financial market risk. Fourier neural operators (FNOs) have been shown to be efficient in solving partial differential equations (PDEs). In this work we demonstrate simulation of dynamics in the chaotic regime of the two-dimensional ($2d$) Kuramoto-Sivashinsky equation using FNOs. Particularly, we analyze the effect of Fourier mode cutoff on the results obtained by using FNOs vs those obtained using traditional PDE solvers. We compare the outputs using metrics such as the $2d$ power spectrum and the radial power spectrum. In addition we propose the normalised error power spectrum which measures the percentage error in the FNO model outputs. We conclude that FNOs capture the dynamics in the chaotic regime of the $2d$ K-S equation, provided the Fourier mode cutoff is kept sufficiently high.

\end{abstract}

\section{Introduction}
Solving Partial Differential Equations (PDEs) is one of the most important and challenging problems in scientific computing. PDEs are used to model a wide range of physical phenomena occurring in scientific and industrial problems. Traditionally, PDEs arising from complex industrial problems are solved on High-Performing Computers (HPC) based on numerical methods \cite{press1992numerical} such as Finite Difference Methods (FDM), Finite Element Method (FEM), Spectral Methods etc. In spite of having high accuracy, these methods require significantly high computational costs and a long time to converge to an accurate solution, prompting researchers to explore novel alternatives that are more efficient and faster.   

Interestingly, Deep Learning (DL), and particularly Deep Neural Networks (DNNs) are emerging as a radically new approach to solve PDEs. DNNs, being a type of universal function approximator, are well known to efficiently handle high-dimensional complexity, making them a promising tool for PDEs \cite{sirignano2018dgm}, \cite{huang2022partial}. By default DNNs admit parallelization of computations, making them easily trainable and infer-able on GPUs. Therefore solving PDEs with them, can straight-away deployed on GPUs, without requiring much mathematical trickery as compared to traditional solvers. Broadly, Deep Learning techniques for PDEs can be divided into two approaches - physics based and data-driven. 

\subsection{Physics-Informed Neural Networks}
Physics-based (commonly referred to as, Physics-Informed Neural Networks (PINNs)) approach, involves integrating underlying physical laws described by PDEs directly into a DNN's training process \cite{raissi2019physics}, \cite{lu2021deepxde}, \cite{cuomo2022scientific}. The key idea, is to approximate the solution of PDE $\vec{u}(\vec{x}, t)$ as a DNN, and train the network with a total loss given by, 
\begin{equation}
\mathcal{L} = \mathcal{L}_{data} + \mathcal{L}_{PDE} + \mathcal{L}_{BC} + \mathcal{L}_{IC}
\end{equation}
where, $\mathcal{L}_{data}$ are few sample data collected from traditional solvers, and the remaining components are self-supervising loss terms. $\mathcal{L}_{PDE}$ ensures that the DNN solution satisfies the governing equation by using Automatic differentiation. $\mathcal{L}_{IC}$, $\mathcal{L}_{BC}$ ensures that the network respects initial and boundary conditions respectively. The pros of PINNs are that they can approximate high-dimensional PDEs, are mesh-free and more efficient to train with limited or noisy training data. However, there are various limitations of PINNs - they require an explicit way of incorporating the PDE and modeling of complex initial and boundary conditions. Also, for a different initial/boundary condition, PINNs needs to be retrained all over again!

\subsection{Data-Driven Approaches}
To avoid explicitly modeling the physics, recently, multiple works have emerged which solve PDEs by a purely data-driven approach. These solvers utilize dynamical data obtained from conventional solvers for training. They then train neural networks to model the dynamics as a transformation between infinite-dimensional function spaces, eliminating the need for explicit physical laws.
The class of such approaches are called Neural Operators, and few examples are Deep Operator Networks (DeepONet) \cite{lu2019deeponet} and Fourier Neural Operators (FNO) \cite{li2021fourierneuraloperatorparametric}. 

DeepONet is a DNN designed to learn non-linear operators. It consists of two components : branch net and trunk net. The branch net processes the input function at discrete spatial and temporal values and transforms it into feature vectors. Concurrently, the trunk net takes the input spatio-temporal coordinates , and transforms them into feature vectors of same dimensions as branch net's feature vectors. These two feature vectors are merged by dot-product and further processed in DNN layers to predict the value at the inputted spatio-temporal coordinates. DeepONets are highly flexible and versatile and perform well for different initial and boundary conditions. They are particularly effective in reducing the generalization errors compared to PINNs. Despite their strengths, DeepONets cannot guarantee overall physics knowledge, involve complex architecture making it difficult to train and struggle at scaling to higher dimensional input functions.

\subsection{Fourier Neural Operators (FNOs)}
To overcome limitations of previously mentioned PDE solvers, \cite{li2021fourierneuraloperatorparametric} came up with a novel approach. FNO is a class of neural operators, which learns the mapping between infinite-dimensional function spaces as a sequence of integral operators acting directly on the Fourier space. FNOs map the input function $a(x)$ to a solution $u(x)$ using a parameterized neural network $G_{\phi}$. The form of $G_{\phi}$ is given by, 
\begin{equation}
G_{\phi} = \mathcal{P} \circ \sigma (W_n + \hat{K}_n ) \circ \dots \circ \sigma (W_1 + \hat{K}_1) \circ \mathcal{Q} a(x)
\end{equation}
here, $\mathcal{P}, \mathcal{Q}$ are lifting operators to higher dimensions and $W_i$ is local linear transformation operator at the $i^{th}$ layer. $\hat{K}_i$, is the non-linear operator given by, 
\begin{equation}
\hat{K} = \mathcal{F}^{-1} \circ R \circ \mathcal{F}  
\end{equation}
where $\mathcal{F}$ represents a non-local Fourier transformation and $R$ is a linear transformation. $\sigma$ is a non-linear activation function which is typically used in PINNs. FNOs have proven to be exceptionally efficient in capturing turbulent flows, physics on complex geometries \cite{li2023fourier}, \cite{li2024geometry}, weather modeling \cite{pathak2022fourcastnet} and many more applications.
Motivated by the success of FNOs, in this work we undertake a numerical study of FNOs for a dynamical system which is well known to exhibit chaos. 

\subsection{Chaotic dynamics and Kuramoto-Sivashinsky equation}
Chaos in PDEs arises when the equations governing the system exhibit non-linearities that lead to complex, unpredictable behavior. These non-linearities can arise from various physical phenomena, such as fluid turbulence, nonlinear optics, and chemical reactions. When these non-linearities are sufficiently strong, the system can exhibit chaotic behavior, characterized by extreme sensitivity to initial conditions, irregular patterns, and the inability to predict future states with certainty. This chaotic behavior can make it challenging to analyze and understand the dynamics of complex systems, as small perturbations can lead to drastically different outcomes. Despite the challenges, studying such systems has great societal impact, particularly to predict weather extremes \cite{blanchard2022multiscaledeeplearningframework}.

The Kuramoto-Sivashinsky (K-S) equation is a nonlinear PDE which exhibits chaos. It has applications in various fields including fluid dynamics, combustion theory, and materials science. It is a canonical model for studying pattern formation and spatio-temporal chaos. In fluid dynamics, it describes the evolution of thin films and interfaces, while in combustion theory, it models the propagation of flames. In materials science, it is used to study the formation of surface patterns and the dynamics of thin films. Additionally, the K-S equation has found applications in finance, where it can be used to model the dynamics of asset prices and the emergence of market bubbles. Due to its versatility and ability to capture complex nonlinear phenomena, the K-S equation remains an active area of research in various scientific disciplines.

FNOs have been used to study the chaotic regime in the two-dimensional ($2d$) Kolmogrov equation \cite{li2023physicsinformedneuraloperatorlearning} and one-dimensional K-S equation \cite{lippe2023pderefinerachievingaccuratelong}. In this paper, we study the $2d$ K-S equation in a square box with Dirichlet boundary conditions,
\begin{equation} \label{eq:K-S}
    \partial_t u = -\frac12 |\nabla u|^2 - \nabla^2 u - \nabla^4 u
\end{equation}
where $u(x,y,t)$ is a scalar field. Here the dimenionality $1d$ vs $2d$ refers to spatial dimensions.


\section{Methodology}
\label{methodology}

In this section, we describe the details regarding generation of the data which is used for training and testing the FNOs, as well as the model architecture.

\subsection{Data generation}
The training data is generated by solving the K-S equation given in equation (\ref{eq:K-S}) using finite difference method for a unit spatial grid of size $128 \times 128$ ($x,y \in (0,128)$) with Dirichlet boundary conditions.
The PDE is evolved up-to time $t = 10$ with time step $\delta t = 0.01$. A dataset of 128 samples is generated where the initial state of the scalar field $u_0(x,y)$ is populated with random values drawn from a uniform distribution with values between $0$ and $1$. Data obtained thus is referred to as \textit{ground truth} in this paper.

Furthermore, on performing spectral analysis on the data generated, we observe a continuous distribution of frequencies which suggests a lack of periodicity or a complex, non-periodic structure, characteristic of chaotic systems (figure \ref{fig:GT_power_spectrum}). 

\subsection{Fourier modes truncation in FNOs} \label{sec:FNO_truncation}
\begin{figure}
  \centering
   \includegraphics[scale = 0.5]{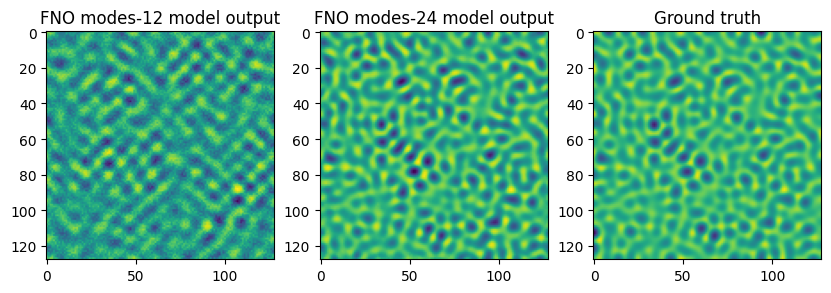}
  \caption{A comparison of output from FNO models vs ground truth.}
  \label{fig:FNO_output}
\end{figure}

In this sub-section we mention details regarding the architecture of the neural network used. In addition to the initial and final fully connected layers, the neural network is made up of 4 FNO layers between these fully connected layers. Since we analyse the effect of the Fourier mode cutoff on the resulting model output, we compare the following two FNO models:
\begin{itemize}
    \item \textit{FNO modes-12} with frequency modes: 12 and hidden channels: 64,
    \item \textit{FNO modes-24} with frequency modes: 24 and hidden channels: 64.
\end{itemize}
The difference in frequency modes in the above two models led to a difference in number of parameters to be trained namely 4,743,937 weights for FNO modes-12 vs 18,899,713 for FNO modes-24. The dataset of 128 samples was divided into train, validation and test datasets of 80:20:20 samples \footnote{Code is available at \href{https://github.com/skhetrapal/chaotic-pde-fno}{github.com/skhetrapal/chaotic-pde-fno}}. Further details regarding the training of the FNO models are mentioned in appendix \ref{app:FNO_train}. The output of the two FNO models is compared to the ground truth for a particular sample in the test dataset in figure \ref{fig:FNO_output}.

\section{Experimental Results}
In order to compare the output of the FNO models with the ground truth, in this section we perform spectral analysis. Particularly, we compare the logarithm of the $2d$ power spectrum and the radial power spectrum of the prediction error between FNO outputs and ground truth.

\subsection{$2d$ Power Spectrum}
\begin{figure}
  \centering
   \includegraphics[width=\linewidth]{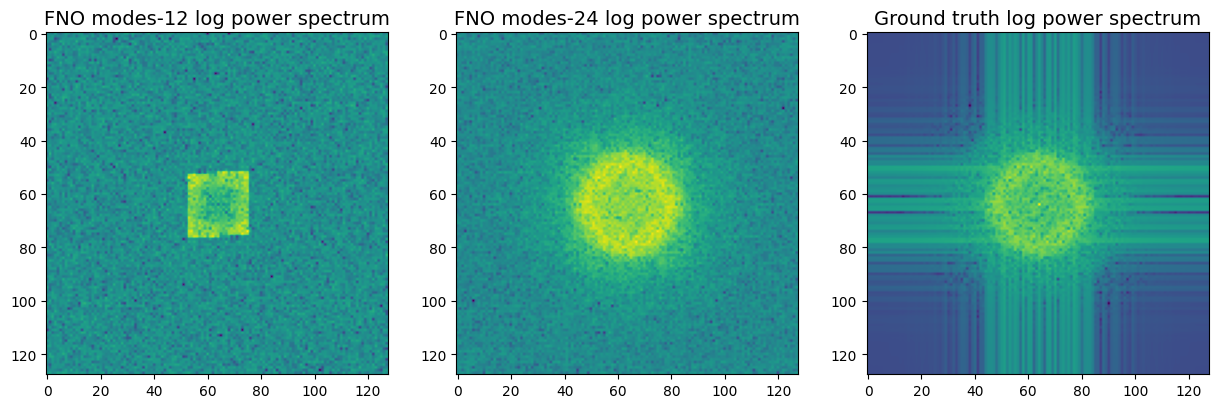}
  \caption{The two-dimensional log power spectrum of FNO model outputs vs ground truth.}
  \label{fig:2d_power}
\end{figure}

The $2d$ power spectrum is a powerful tool for analyzing the frequency content of $2d$ data. It provides a visual representation of the energy distribution in the frequency domain and helps identify dominant patterns. In order to obtain this we first perform the $2d$ Fourier transform:
\begin{equation} \label{eq:fourier_transform}
    F(k_x, k_y) = \int_{-\infty}^{\infty} \int_{-\infty}^{\infty} f(x, y) e^{-i2\pi(k_x x + k_y y)} dx dy .
\end{equation}
The power spectrum is obtained by taking the square of the magnitude of the Fourier transform:
\begin{equation}
    P(k_x, k_y) = |F(k_x, k_y)|^2 =\mathrm{Re}(k_x, k_y)^2 + \mathrm{Im}(k_x, k_y)^2.
\end{equation}
We compare the logarithm of this power spectrum $\log P(k_x, k_y)$ for FNO modes-12 and modes-24
with the ground truth in figure \ref{fig:2d_power}. The component of the Fourier transform with zero frequency (DC component) is at the centre of the matrix. We see that the FNO model with Fourier modes cutoff at 24 captures a lot more spectral features of the ground truth compared to the FNO model with Fourier modes cutoff at 24.

\subsection{Radial power spectrum for prediction error}

\begin{figure}
  \centering
  \includegraphics[width=0.495\linewidth]{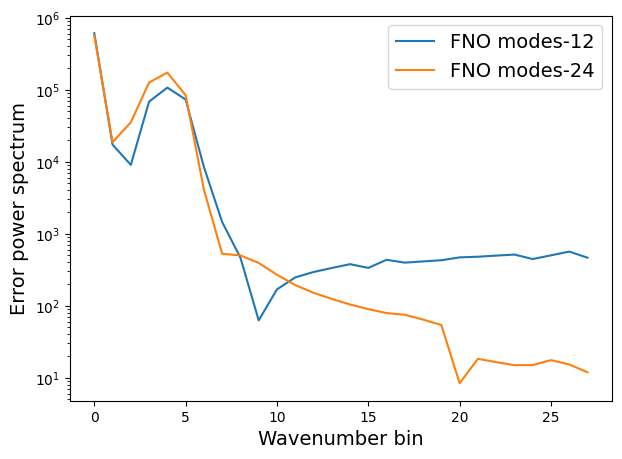}
   \includegraphics[width=0.495\linewidth]{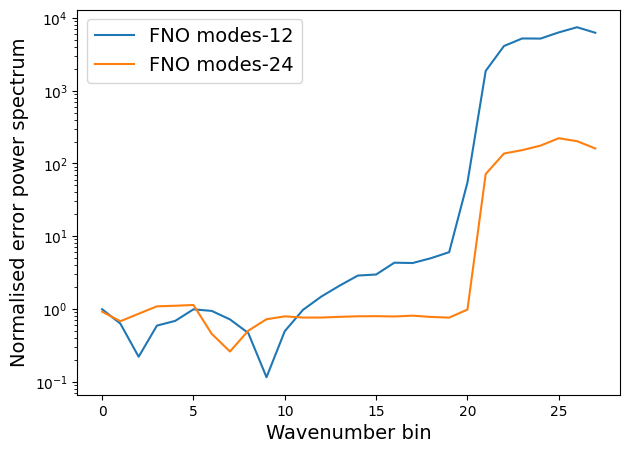}
  \caption{A comparison of error power spectrum (left) and normalised error power spectrum (right) vs wavenumber for FNO modes-12 and FNO modes-24.}
  \label{fig:error_power}
\end{figure}

The radial power spectrum provides a clear visualization of the energy distribution in the frequency domain, focusing on the radial variation rather than the $2d$ grid. Thus, it is useful in analyzing patterns and features in the frequency content of the output. The steps for computing the same are given below:

\begin{enumerate}
    \item Obtain the frequency-domain representation of the output of the FNO model by performing the $2d$ fast Fourier transform (FFT) given in equation (\ref{eq:fourier_transform}). 
    \item Determine the radial wave number for each entry in the Fourier-transformed matrix, which is the distance from the center of the matrix $P(k_x,k_y)$.
    \item Average the power spectrum along the radial wavenumbers. This is done by binning the spectrum elements based on their radial wavenumber and calculating the average power within each bin. As an example, for the $128 \times 128$ output, we use $28$ bins.
\end{enumerate}
We call the results obtained by following these steps: \textit{radial power FNO (modes-12/modes-24) output} and \textit{radial power ground truth}. Next, the absolute difference between these two is computed, shown in figure \ref{fig:error_power} (left). We find that at higher wavenumber, the energy in prediction error in case of FNO modes-24 model is lower compared to FNO modes-12 model.

While the radial power spectrum for prediction error is a  useful tool used in literature \cite{qin2024betterunderstandingfourierneural, lippe2023pderefinerachievingaccuratelong}, it fails to provide a clear visualization of percentage error in the radial power spectrum of the FNO model outputs vs ground truth. In order to take this into account, we propose the following metric to analyze the error:
\begin{equation}
    \textrm{Normalised error power spectrum} = \left|\frac{\textrm{radial power FNO output} - \textrm{radial power ground truth}}{\textrm{radial power ground truth}}\right|
\end{equation}
The plot of the normalised error power spectrum vs radial wavenumber is shown in figure \ref{fig:error_power} (right). We find that the percentage error in the radial power spectrum remains similar at small wavenumber for both the cases, i.~e.~ Fourier modes cutoff 12 and 24. However, there is an exponential increase in power for higher wavenumbers for FNO modes-12 at wavenumber bin 10 compared to at wavenumber bin 20 for FNO modes-24. Also, the energy in the latter remains lower at the highest wavenumbers.

\section{Conclusion}

In this work we demonstrate that FNOs can capture the dynamics in the chaotic regime of the $2d$ K-S equation, given that Fourier mode cutoff is kept sufficiently high. We compare the performance of two FNO models where Fourier mode cutoff is kept at 12 and 24 respectively. On comparing the $2d$ power spectrum, radial power spectrum of prediction error and normalised error power spectrum, we find that the latter model, i.~e.~ FNO modes-24 captures the chaotic dynamics of the ground truth better upto to higher wavenumbers. This is also due to the higher number of parameters trained in FNO modes-24 compared to FNO modes-12. However, even for FNO modes-24, the training loss and validation loss achieved are 0.13677 and 0.29458, respectively. This suggests that the model can get better convergence with more training data.

\subsection{Future Work}
While the study has shown that FNOs can simulate the temporal dynamics of the chaotic KS equation, it would be interesting to see if it can also capture the higher order statistical properties of chaos in the system. More specifically, measuring the Lyapunov exponent in this system and bench-marking against traditional solvers will provide us a better understanding of how well FNOs perform in this setup \cite{edson2019lyapunov}.  

In a recent work, it has been demonstrated that FNOs can learn dynamics in quantum spin systems \cite{shah2024fourierneuraloperatorslearning}. While the systems studied in this work are integrable, it would be interesting to see how efficient are FNOs in learning dynamics of quantum chaotic systems \cite{Alba:2019ybw, khetrapal2024mutualinformationscramblingising}.

\section*{Acknowledgements}
SK’s research is supported by Department of Science and Technology’s INSPIRE grant DST/INSPIRE/04/2020/001063.

\bibliography{references}  

\appendix

\section{Appendix}
\subsection{Spectral analysis of training data}
\begin{figure}
  \centering
   \includegraphics[width=0.495\linewidth]{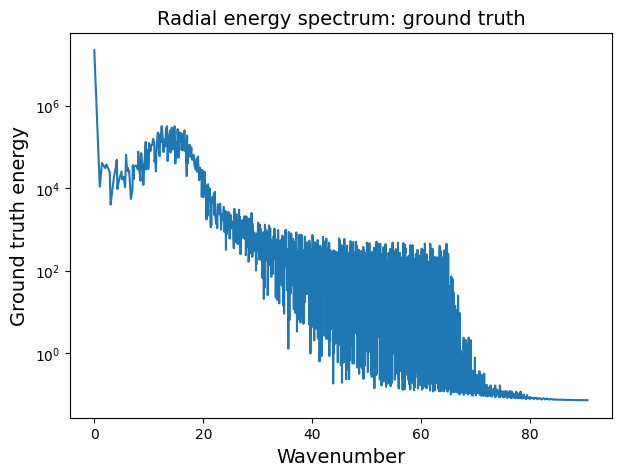}
   \includegraphics[width=0.495\linewidth]{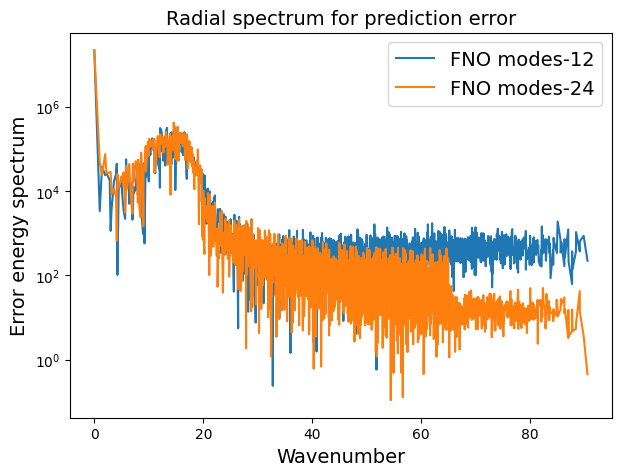}
  \caption{Radial power spectrum of ground truth vs radial wavenumber (left) and power spectrum of error between FNO model outputs and ground truth vs wavenumber (right).}
  \label{fig:GT_power_spectrum}
\end{figure}

The radial power spectrum of ground truth result shows a continuous distribution of frequencies, i.~e.~a broadband spectrum as seen in figure \ref{fig:GT_power_spectrum} (left) which implies chaotic behavior. This is because chaotic systems tend to exhibit complex, non-periodic patterns that lack a clear, repeating structure. Conversely, a discrete spectrum, featuring distinct, sharp peaks, suggests more regular behavior. Such peaks  correspond to periodic or quasi-periodic components, implying a predictable or cyclical pattern.

\subsection{Details of FNO model training} \label{app:FNO_train}
\begin{figure}
    \centering
    \includegraphics[width=0.495\linewidth]{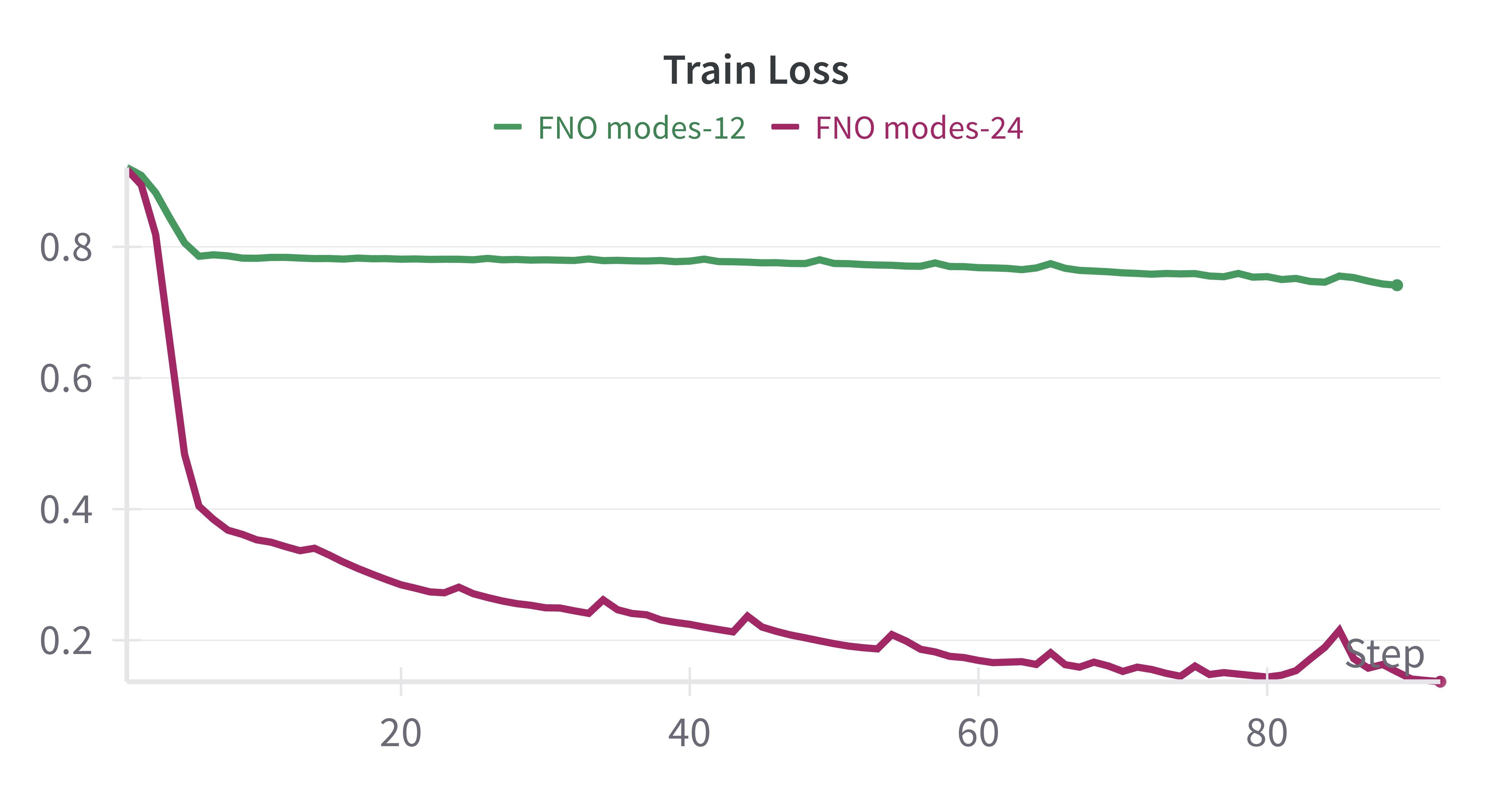}
    \includegraphics[width=0.495\linewidth]{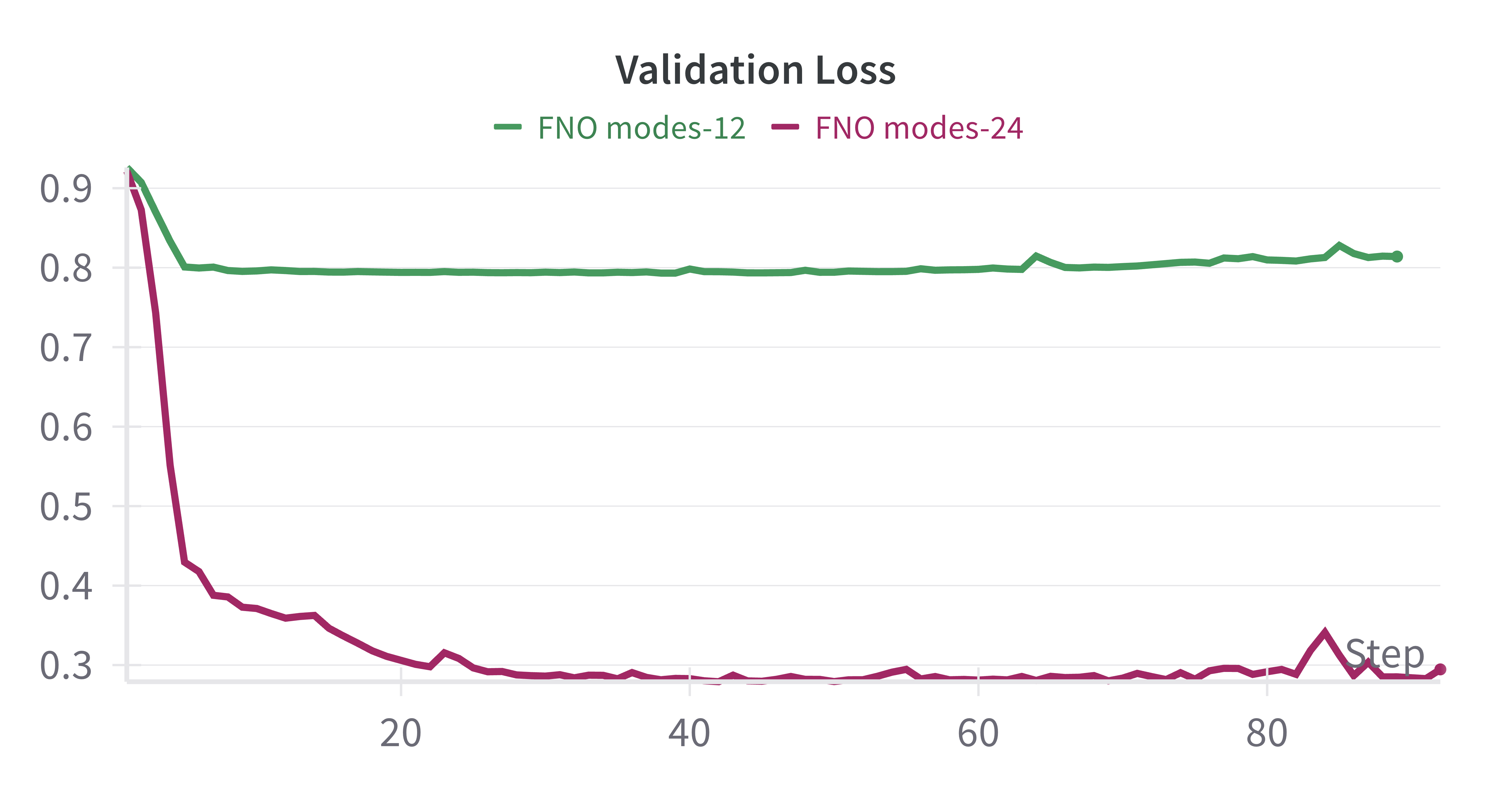}
    \caption{A comparison of training and validation loss of the two FNO models}
    \label{fig:training_loss}
\end{figure}

This appendix provides details regarding the training of the two FNO models mentioned in section \ref{sec:FNO_truncation}. 
The hyper-parameters used in experimental study of FNOs are the following: 
\begin{multicols}{2}
\begin{itemize}
\item Loss function: relative L$_2$ loss
\item Optimizer: Adam
\item Scheduler: step learning rate 
\item Test loss metric:	MSE
\item Weight decay:	0.0001
\end{itemize}
\end{multicols}
The FNO modes-12 model was trained for 89 epochs and FNO modes-24 model was trained for 92 epochs. Due to difference in trainable parameters viz.~ 4,743,937 for the former vs 18,899,713 for the latter, this led to a vast difference in training and validation loss achieved in the training of these two model as shown in figure \ref{fig:training_loss}.

\end{document}